\documentclass[runningheads]{llncs}

\usepackage[year=2026]{eccv}

\usepackage{eccvabbrv}   % \eg \ie \etc \cf \etal
\usepackage{microtype}
\microtypesetup{expansion=true,protrusion=true,factor=1200,stretch=30,shrink=30}
\usepackage{graphicx}
\usepackage{booktabs}
\graphicspath{{figures/}}
\usepackage[accsupp]{axessibility}

\usepackage{hyperref}
\usepackage{orcidlink}

\begin{document}

% ---------------------------------------------------------------
\title{SCOUT: Sim-to-Real Text-Based Person Retrieval by Embedding-Space Prediction over Frozen Video Features}

\titlerunning{SCOUT: Frozen Embedding-Space Prediction for Person Retrieval}

\author{Abdarahmane Traor\'e\orcidlink{0000-0002-8942-4349} \and
Andy Couturier\orcidlink{0000-0002-3854-544X} \and
\'Eric Hervet\orcidlink{0000-0002-3117-6506}}
\authorrunning{A. Traor\'e et al.}
\institute{Embia, Computer Science Department, Faculty of Science,\\
Universit\'e de Moncton, Moncton, NB, Canada\\
\email{\{eat4651, andy.couturier, eric.hervet\}@umoncton.ca}}

\maketitle

\begin{abstract}
Text-based person retrieval under a sim-to-real gap (synthetic training data, a real-image
gallery) is typically tackled with expensive fine-tuned cross-encoders. We ask whether a
\emph{frozen-encoder} system can compete. We present \textbf{SCOUT}, which casts cross-modal
retrieval as \emph{prediction in embedding space}. A trainable predictor maps the patch tokens of a
frozen video encoder into the embedding space of a frozen text encoder under a bidirectional InfoNCE
objective, and no encoder is fine-tuned in the base model. The video encoder is V-JEPA, the text
encoder is EmbeddingGemma, and the predictor is initialized from a Qwen3.5-0.8B decoder. We make
three findings. First, the best frozen text encoder is simply the one whose
geometry best matches the video features. A training-free alignment score ranks three candidate
text encoders in the same order as their retrieval accuracy on our held-out split (Spearman
$\rho = 1.0$); a fourth, LLM-based encoder shows the rule is metric-dependent, holding for a
neighborhood-overlap score ($\rho = 0.8$) but not for a linear probe ($\rho = -0.2$).
Second, two precision-targeted levers, parameter-efficient ExPLoRA adaptation of the video encoder
and a training-free attribute-decomposed reranker built on a vision-language model (VLM), improve
the top-rank precision that
otherwise limits the frozen system, together adding 2.2 points of leaderboard rank-1 recall (R@1).
Third, a
local-versus-public calibration study explains which interventions transfer to the real domain. On
AI City Challenge 2026 Track~4 the full retrieve-fuse-rerank system reaches 84.25 mAP@10 on the
final leaderboard, while a single frozen model submitted alone reaches 60.63. Our trained components
cost about 95 GPU-hours in total. CMP (cross-modal pose-aware), the dataset authors' fine-tuned
cross-encoder that trains for sixteen GPU-days, is also one fusion member of the full leaderboard
system, not an alternative it avoids. Code and annotations are available at
\url{https://github.com/abtraore/SCOUT-ECCV}.
\end{abstract}

\section{Introduction}
\label{sec:intro}

Text-based person retrieval asks a system to return, from a large image gallery, the person who
matches a free-form natural-language description of their \emph{appearance}, \emph{action}, and
\emph{scene}~\cite{aicity26,Tang26AICity26}. The sim-to-real variant studied in AI City Challenge 2026 Track~4,
officially \emph{Text-Based Person Re-Identification (Sim2Real)}, sharpens the problem. The training
data are diffusion-generated synthetic images, while the test gallery is real photographs, so a model
must generalize across a domain gap with no real labels. Performance is measured by mean average precision over the
top-10 retrieved images (mAP@10), with exactly one correct gallery image per query.

The dominant high-performance approach to this task fine-tunes models of the ALBEF/X-VLM
family~\cite{albef,xvlm}, built around an image-text \emph{cross-encoder}, a fusion module that
jointly attends over a (caption, image) pair and emits a match score. They are powerful but
expensive in two ways: training fine-tunes the full backbone (the dataset authors' model trains for
sixteen GPU-days~\cite{cmp}), and a cross-encoder cannot serve as a first-stage retriever, since
scoring an $N$-image gallery costs $N$ fusion forward passes rather than $N$ cached embeddings.
These systems therefore retrieve with a dual-encoder stage first and rerank only a shortlist.

We ask whether a \emph{frozen-encoder} system can compete. Our starting point is the idea, central
to Joint-Embedding Predictive Architectures (JEPAs)~\cite{ijepa,vjepa}, that representations can be
learned by \emph{predicting in embedding space} rather than reconstructing pixels or generating
tokens. Its vision-language form, VL-JEPA~\cite{vljepa}, predicts a caption embedding from video,
and SCOUT is closest to it. We adapt that recipe to a frozen-encoder retriever. A self-supervised
\emph{video} encoder (V-JEPA) and a text encoder stay entirely frozen. Only the predictor is
trained. It maps the video encoder's patch tokens into the frozen text embedding space, under a
bidirectional InfoNCE objective~\cite{infonce}. VL-JEPA instead uses a regression objective and a
trainable text projection. Neither encoder is contrastively co-trained, as CLIP's are~\cite{clip}.
The result is a bi-encoder rather than a cross-encoder, so gallery embeddings are computed once and
queried with a dot product. Cross-encoding enters our system only later, as an optional reranker
over a short candidate list (\cref{sec:systemeval}). We call the method \textbf{SCOUT}.

This design raises three questions that organize the paper. \emph{(i) What is the right frozen text
target?} Because the predictor maps \emph{into} a fixed text space, that geometry is decisive, and a
training-free alignment score selects it. \emph{(ii) Can a frozen system reach the top-rank
precision of a fine-tuned one?} The gap to the top teams on the challenge leaderboard is almost
entirely rank-1 recall (R@1), a reranking-quality problem rather than one of candidate recall, which
two precision-targeted levers address. \emph{(iii) Which improvements transfer to the real domain?}
A local-versus-public calibration study answers it, with a decorrelation statistic that predicts
which additions help.\looseness=-1

Our contributions are: \textbf{(1)} SCOUT, a frozen-encoder architecture that casts text-to-image
retrieval as embedding-space prediction; \textbf{(2)} a training-free alignment \emph{heuristic}
for screening the frozen text target, a neighborhood-overlap score that stays predictive against a
falsifier where a linear-probe version of the same idea does not; and \textbf{(3)} two precision levers that
improve R@1 beyond the frozen-encoder base, adapting the video encoder and decomposing the VLM
reranker, together adding 2.2 points of leaderboard R@1 to
the full system, along with the calibration methodology that justifies them. We also map where
further gains stop.\looseness=-1

\section{Related Work}
\label{sec:related}

\subsubsection{Text-based person retrieval and cross-encoders.}
The dominant approach to text-image person retrieval fine-tunes models of the ALBEF/X-VLM
family~\cite{albef,xvlm}, in which a contrastive dual-encoder stage shortlists candidates and a
\emph{cross-encoder} that co-attends over each (caption, image) pair reranks them. The Track~4
dataset's own baseline, CMP~\cite{cmp}, is such a model, fine-tuned on the full training corpus.
SCOUT is the frozen bi-encoder counterpart.

\subsubsection{Embedding-space prediction (JEPA).}
Joint-embedding predictive architectures learn to predict masked latents in representation space
rather than reconstructing inputs, with I-JEPA for images~\cite{ijepa} and V-JEPA for
video~\cite{vjepa1,vjepa}.
The vision-language JEPA, VL-JEPA~\cite{vljepa}, carries the idea cross-modally, predicting a text
embedding from video, and SCOUT is closest to it. SCOUT differs from it in two ways that later sections
show matter: a bidirectional InfoNCE objective in place of VL-JEPA's regression-plus-regularization
(\cref{sec:method}), and a fully frozen text target with no learned projection into it
(\cref{sec:alignment}).

\subsubsection{Frozen-feature transfer and gentle adaptation.}
When a head is already trained on frozen features, full fine-tuning distorts the pretrained
representation and underperforms out of distribution (OOD), a finding by Kumar~\etal~\cite{lpft},
who propose linear-probe-then-fine-tune (LP-FT) in response. Parameter-efficient methods adapt a
backbone with few trainable parameters, such as
low-rank adaptation (LoRA~\cite{lora}) and its extended-pretraining form ExPLoRA~\cite{explora}. Our
\cref{sec:explora} recipe combines a measured block-unfreeze window with LoRA and a
WiSE-FT-style~\cite{wiseft} zero initialization.\looseness=-1

\subsubsection{Representation alignment and fusion.}
The Platonic-representation hypothesis~\cite{platonic} argues that strong models converge to aligned
representations, measurable without training via mutual-$k$NN overlap or centered kernel alignment
(CKA). We use such measures to
\emph{select} SCOUT's frozen target (\cref{sec:alignment}). On the system side we combine retrievers
by CombSUM score fusion~\cite{combsum}, deliberately score-level rather than rank fusion~\cite{rrf}
(which our leaderboard study finds regresses, \cref{sec:calib}), and rerank a short list with a
vision-language model (VLM) cross-encoder~\cite{qwen3vl}.

\section{SCOUT: Prediction in Embedding Space}
\label{sec:method}

\begin{figure}[tb]
  \centering
  \includegraphics[width=\linewidth]{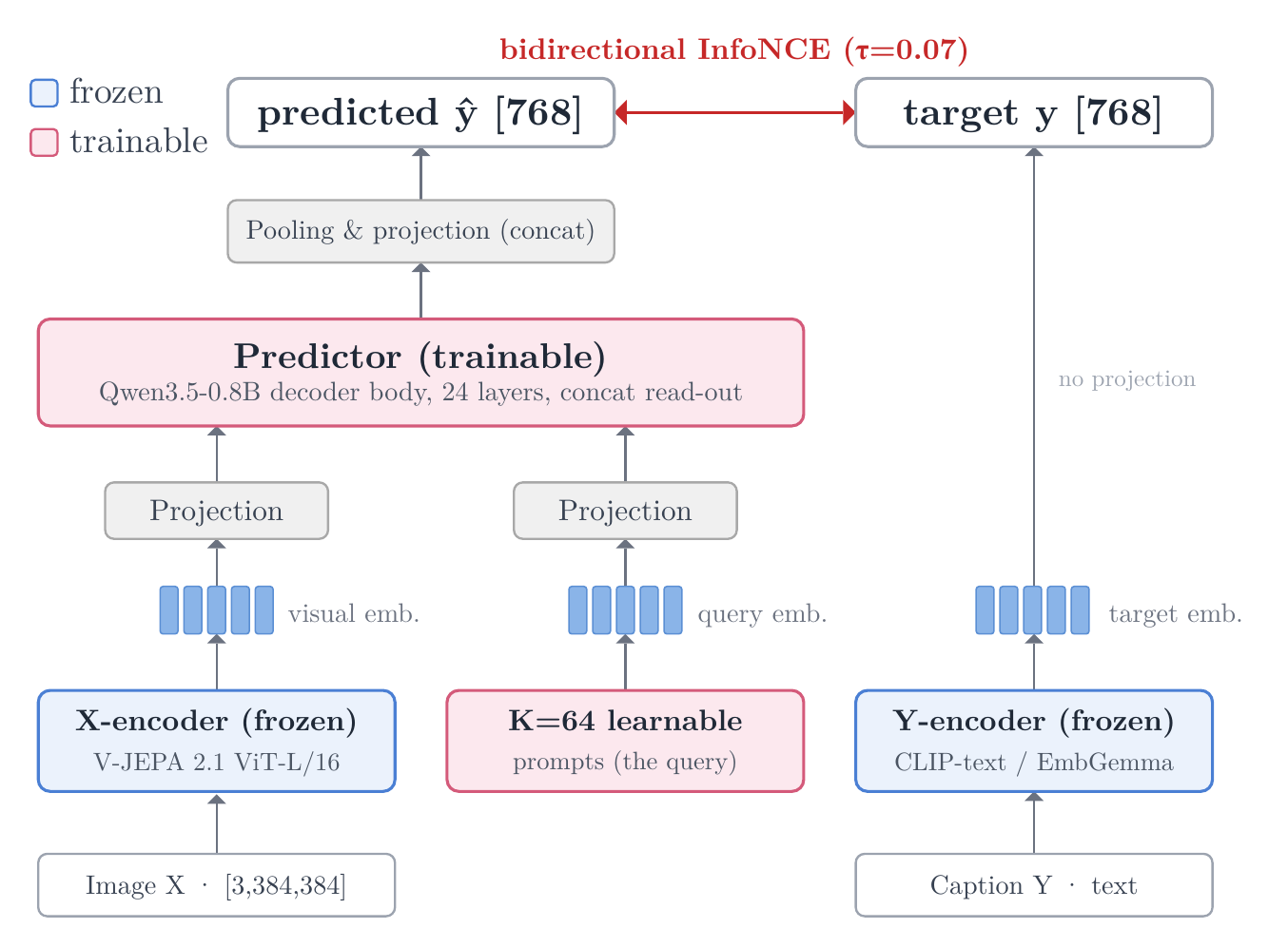}
  \caption{SCOUT. A frozen self-supervised video encoder (V-JEPA) produces patch tokens, and a trainable
  predictor with $K$ learnable prompt tokens reads them out into the embedding space of a
  \emph{frozen} text encoder. Training is a bidirectional InfoNCE between the predicted and the true
  caption embedding, and at inference the gallery is encoded once (bi-encoder).\looseness=-1}
  \label{fig:arch}
\end{figure}

\subsection{Task and Notation}
\label{sec:notation}
A query is a caption, and the gallery is a set of images $\{x_n\}_{n=1}^{N}$. Retrieval ranks the
gallery by similarity to the caption in a shared $d$-dimensional space ($d = 768$). Each query has
exactly one relevant gallery image, so average precision reduces to the reciprocal rank of that image
and mAP@10 rewards placing it at rank one.

\subsection{Architecture}
\label{sec:arch}
SCOUT has three parts (\cref{fig:arch}), and only the predictor is trained in the base model.\looseness=-1

\textbf{Frozen video encoder (X).} We use a V-JEPA 2.1 ViT-L/16~\cite{vjepa}, kept frozen, as a
generic spatial feature extractor over a single frame. It emits $P$ patch tokens
$X = (x^1, \dots, x^P)$. V-JEPA was itself pretrained by predicting masked latents in embedding
space, a natural substrate for our predictor.

\textbf{Frozen text encoder (Y).} A frozen text encoder maps the caption to a unit vector
$y \in \mathbb{R}^d$. It is never trained, and \cref{sec:alignment} shows its geometry is the
decisive design choice.\looseness=-1

\textbf{Predictor.} A trainable transformer body of 24 layers, initialized from a Qwen3.5-0.8B
decoder~\cite{qwen}, receives the projected patch tokens followed by $K$ learnable \emph{prompt
tokens} $(p^1, \dots, p^K)$. After self-attention, the $K$ prompt outputs are \emph{concatenated}
and linearly projected to a unit vector $\hat{y} \in \mathbb{R}^d$, the predicted caption embedding.
The concat read-out gives each prompt its own output slot, which we find scales far better than
mean-pooling the prompts (\cref{sec:ablation}).

\subsection{Objective}
\label{sec:objective}
We train with a symmetric in-batch InfoNCE~\cite{infonce}:
\begin{equation}
\mathcal{L} = -\frac{1}{2B} \sum_{i=1}^{B} \left[
  \log \frac{e^{s(\hat{y}_i, y_i)/\tau}}{\sum_{j=1}^{B} e^{s(\hat{y}_i, y_j)/\tau}}
  + \log \frac{e^{s(\hat{y}_i, y_i)/\tau}}{\sum_{j=1}^{B} e^{s(\hat{y}_j, y_i)/\tau}}
\right],
\label{eq:infonce}
\end{equation}
where
\begin{itemize}
  \setlength{\itemsep}{0pt}
  \setlength{\parskip}{0pt}
  \item $B$ is the number of (image, caption) pairs in one forward pass, indexed by $i$ and $j$;
  \item $\hat{y}_i$ is the \emph{predicted} caption embedding of image $i$, the predictor output of
        \cref{sec:arch};
  \item $y_j$ is the frozen text encoder's embedding of caption $j$;
  \item all embeddings are $L_2$-normalized, so the score $s(a,b) = a^\top b$ is their cosine
        similarity, in $[-1, 1]$;
  \item $\tau$ is the softmax temperature, set to $\tau = 0.07$.
\end{itemize}
Each fraction in \cref{eq:infonce} is a softmax over the batch. In the first term the prediction
$\hat{y}_i$ must pick out its own caption $y_i$ among all $B$ captions (the image-to-text direction).
In the second, the caption $y_i$ must pick out its own prediction among all $B$ predictions
(text-to-image). The matched pair $(\hat{y}_i, y_i)$ is the positive of both terms, the other $B-1$
candidates are the in-batch negatives, and the $1/2B$ prefactor averages the two directions over the
batch.

The quantity in \cref{eq:infonce} that matters most in practice is $B$ itself. The contrastive
signal depends on the \emph{per-forward} batch, not the effective batch under gradient accumulation.
Under data-parallel training $B$ is moreover the \emph{per-GPU} batch: each rank evaluates the
softmax over its own local batch, so adding GPUs raises throughput but leaves the negative count per
query untouched. Pooling negatives \emph{across} ranks requires a separate mechanism, an explicit
cross-GPU all-gather of the embeddings, as popularized by CLIP~\cite{clip}.
\cref{sec:ablation} shows this pooling \emph{saturates} past the per-rank operating point, and that
the per-rank $B$, which sets both the in-batch negatives and the sample diversity of each gradient
step, is the single most influential training parameter.

\subsection{Inference and Efficiency}
\label{sec:efficiency}
SCOUT is a bi-encoder. Gallery images are passed through V-JEPA and the predictor once to produce
$\hat{y}_n$, a query caption is encoded by the frozen text encoder to $y$, and ranking is by cosine
similarity $s(\hat{y}_n, y)$. Encoding the $36{,}773$-image test gallery takes about a minute on
one GPU, and the embeddings are cached. A query then costs one text-encoder forward and $N$ dot
products, milliseconds in total, versus a cross-encoder's $N$ full forward passes. The one
non-negligible inference cost in the full system of \cref{sec:systemeval} is its VLM reranker,
whose full-test pass ($20$ candidates per query, $39{,}560$ pairs per signal) takes about half an
hour on a four-GPU server.

Efficiency comes from \emph{what is frozen}, not from a small parameter count. The predictor is
sizable, $549.6$M trainable parameters, but training never back-propagates into a frozen backbone,
and the base model converges in about $7.4$ hours on six RTX 5090 GPUs (44 GPU-hours,
peaking near $22$ GB of each card's $32$). The ExPLoRA update of \cref{sec:levers} trains $85.8$M
further encoder-side parameters in 5.2 hours (31.2 GPU-hours, $16.4$ GB per GPU at its batch of
$64$), and
the VLM reranker lever is training-free. With the $29.73$M ScoutITM image-text-matching head
(\cref{sec:itm}), every trained component fits in about 95 GPU-hours. The fully fine-tuned
CMP~\cite{cmp} trains for sixteen GPU-days (384 GPU-hours) on four RTX 3090 GPUs, a
total-training-cost contrast for the components we train rather than a matched per-pass benchmark.
The leaderboard system also reads CMP's own scores as one retrieval-stage member
(\cref{sec:system}), so its 384 GPU-hours are a dependency of that system rather than a cost the
system avoids.

\section{The PAB Benchmark}
\label{sec:dataset}

\begin{figure}[tb]
  \centering
  \includegraphics[width=0.225\linewidth]{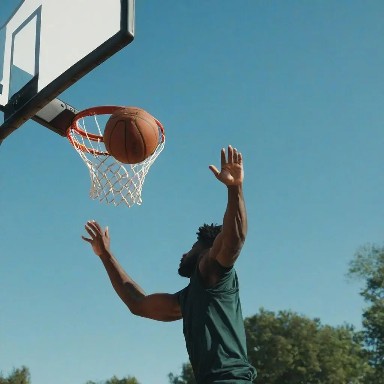}\hfill
  \includegraphics[width=0.225\linewidth]{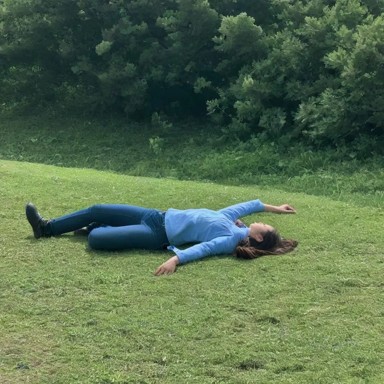}\hfill
  \includegraphics[width=0.225\linewidth]{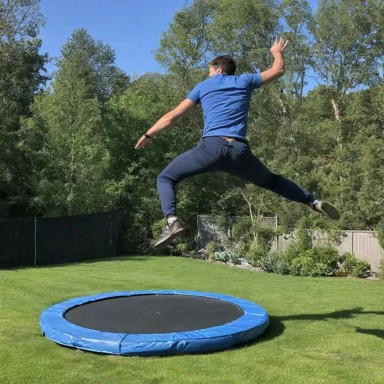}\hfill
  \includegraphics[width=0.225\linewidth]{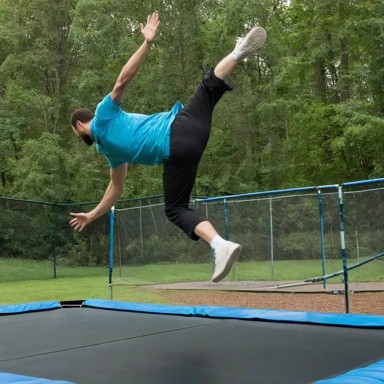}
  \caption{Representative \emph{synthetic} PAB training pairs: a \emph{normal} activity, an
  \emph{anomaly}, and a \emph{hard pair}, the same scene generated once normal and once anomalous,
  captions differing only in action.}
  \label{fig:pab}
\end{figure}

We evaluate on the Pedestrian Anomaly Behavior (PAB) benchmark~\cite{cmp}, the dataset adopted by AI
City Challenge 2026 Track~4~\cite{aicity26}. PAB frames person retrieval as a \emph{sim-to-real}
problem. The training set is 1,013,605 \emph{synthetic} image-caption pairs, generated with the
Realistic Vision V4.0 diffusion model and captioned by the Qwen2-VL multimodal language
model~\cite{cmp}. The test set is 1,978
query captions over a gallery of 36,773 \emph{real} photographs, with exactly one relevant gallery
image per query. Each caption describes a person's \emph{appearance}, \emph{action}, and
\emph{scene}. The test set is balanced one-to-one between \emph{normal} and \emph{anomalous}
behavior (for example playing or performing versus lying or being struck), so the task couples
fine-grained description matching with sensitivity to the anomaly. No real-domain labels are
released, and the resulting synthetic-to-real gap is the central difficulty.
\Cref{fig:pab} shows representative training pairs, including a \emph{hard pair}, two records
that differ only in action.

\section{Empirical Analysis}
\label{sec:analysis}

\textbf{Setup.} On the PAB benchmark (\cref{sec:dataset}) we hold out for offline study a frozen
5,000-record validation split from the synthetic training data, 2,500 per behavior class,
scored against 35,000 random distractors, reproduced deterministically by our released code.
We report mAP@10 and recall at rank $k$ (R@$k$),
the fraction of queries whose one relevant image lands in the top $k$. The in-challenge public
leaderboard scored a 50\% test subset, and after the close every submission was re-scored on the
full test set. \emph{Leaderboard} numbers throughout are these full-test scores, and we note the
subset reading where it shaped a decision. On the final leaderboard (2026-07-10 close) our best
submission scores 84.25 mAP@10 / 75.63 R@1, ranking 18th of 28 teams, while the leader scores
99.30 / 98.74. All mAP and R@$k$ values are reported as percentages. Throughout, \emph{val}
numbers are single-model scores on this split, and \emph{leaderboard} numbers refer to the full
system of \cref{sec:system} unless marked single-model. Unless stated, every SCOUT run shares one recipe:
AdamW (betas $0.9/0.95$), learning rate $2 \times 10^{-5}$ with linear warmup over the first $600$
of $6000$ steps then cosine decay to zero, weight decay $0.01$, $\tau = 0.07$, and a training
split excluding val. Each ablation therefore changes a single factor, from one training run each.
Close comparisons reflect the scale of deltas we observe across the sweep, not a formal
significance test.\looseness=-1

\subsection{Ablation Study}
\label{sec:ablation}
\Cref{tab:ablation} traces the ablation that built SCOUT, each row extending the one above
(the $K$ row aggregates the stepwise growth of $K$ from $1$ to $64$ plus the concat read-out), and
every number in this subsection is val mAP@10. Four lessons stand out. \textbf{(i) Class balance.}
Matching the test set's one-to-one normal-to-anomalous balance during sampling adds $+1.6$ points. \textbf{(ii)
The predictor, not the video encoder, is the bottleneck.} Enlarging the \emph{frozen} video encoder alone
(ViT-B to ViT-L at a shallow predictor) does not significantly change accuracy. Deepening the
predictor from $8$ to $24$
layers adds $+1.9$ points and lets the larger video encoder contribute. \textbf{(iii) Read-out.} $K$ learnable prompt
tokens with a \emph{concat} read-out, where each prompt keeps its own output slot, scale where
mean-pooling collapses. $K$ saturates near $64$, and doubling further to $128$ loses $1.5$ points. \textbf{(iv) Contrastive batch dominates, but the
lever is the per-rank batch, not the raw negative count.} Raising the per-forward InfoNCE batch from
$16$ to $128$ adds $+9.4$ points, the largest single-factor gain in the study. Because every run
trains for the same $6000$ steps, batch $128$ also processes eight times more samples in total, a
confound the ablation does not isolate from batch size itself. The batch-$16$ run's own training
loss argues against sample count alone driving the gap. It falls from $2.74$ at step $0$ to $0.055$
by step $2000$ of $6000$ and barely moves after ($0.039$ at the end), a suggestive sign, not a
controlled test, that its remaining steps bought little. This batch is \emph{per-GPU}
(\cref{sec:objective}).
To isolate which factor pays, we pool negatives \emph{across} all six ranks with a gradient-exact
cross-GPU all-gather, enlarging the pure negative pool from $127$ to $767$ at \emph{identical} per-rank
batch, compute, and samples-per-step. Gathering does not significantly improve the result, $86.79$
mAP@10 with gathering against $86.39$
without at matched eval resolution, within noise (last row of \cref{tab:ablation}). The per-rank
forward is doing the work through more distinct samples and better gradient quality per step, and
the negative count alone saturates already at the per-rank pool of $127$. We select models on
val mAP.

\begin{table}[tb]
  \centering
  \caption{Ablation sequence on the held-out val split (EmbeddingGemma text target), each row adding one
  change to the row above. The last row is a control off the batch-128 model that pools InfoNCE
  negatives across GPUs, evaluated at the $384$-px training resolution (a within-noise $86.39$
  for the batch-128 model there, against $86.57$ at the earlier $256$-px eval).}
  \label{tab:ablation}
  \begin{tabular}{@{}lccccr@{}}
    \toprule
    Change & X & Pred. & $K$ / read-out & Bal. & mAP@10 (\%) \\
    \midrule
    CLIP-L~\cite{clip} (frozen, joint) & ViT-L/14 & n/a & n/a & n/a & 56.89 \\
    SigLIP2-L~\cite{siglip2} (frozen, joint) & ViT-L/16 & n/a & n/a & n/a & 68.95 \\
    base recipe & ViT-B/16 & 8 & 1 / mean & none & 59.35 \\
    + class balance & ViT-B/16 & 8 & 1 / mean & class & 60.99 \\
    + predictor depth 8$\to$24 & ViT-L/16 & 24 & 1 / mean & class & 62.90 \\
    + $K$=64, concat read-out & ViT-L/16 & 24 & 64 / concat & class & 77.19 \\
    + InfoNCE batch 16$\to$128 & ViT-L/16 & 24 & 64 / concat & class & 86.57 \\
    + pre-mined hard pairs & ViT-L/16 & 24 & 64 / concat & class & \textbf{87.90} \\
    \midrule
    gather negs 127$\to$767 (control) & ViT-L/16 & 24 & 64 / concat & class & 86.79 \\
    \bottomrule
  \end{tabular}
\end{table}

\subsection{Frozen-Target Selection: An Alignment Problem}
\label{sec:alignment}
Because SCOUT predicts \emph{into} a fixed text space, the learnability of the map (and thus
retrieval quality) is set by that space's geometry relative to the video features. We make this
quantitative. On the val pairs we measure the alignment between $X$ (frozen V-JEPA patch tokens,
mean-pooled) and each candidate frozen $Y$ with three \emph{training-free} lenses: mutual
$k$-nearest-neighbour overlap (the Platonic-representation measure~\cite{platonic}), linear
CKA~\cite{cka}, and the $R^2$ of a ridge linear probe $X \rightarrow Y$, a linear stand-in for
SCOUT's predictor. \Cref{tab:yencoder} shows that within the CLIP / EmbeddingGemma / PE-Core~\cite{pecore} family,
all three measures order the candidate text encoders \emph{identically} to the val mAP of the
SCOUT model trained against each (Spearman $\rho = 1.0$), with no training. Alignment to V-JEPA can
therefore select a candidate
target before any multi-GPU-hour run. It also explains an otherwise surprising result. CLIP-text~\cite{clip},
at $123.65$M parameters, outperforms the larger text-only EmbeddingGemma~\cite{embeddinggemma}
($300$M). The smaller encoder was trained by an \emph{image-text} loss, so its manifold is already
shaped toward the image side. The choice of frozen $Y$ follows image-alignment rather than parameter
count or text-only retrieval quality.

This criterion holds within a family of comparable encoders, and a falsifier marks its limit.
The large-language-model (LLM) embedding encoder
Qwen3-Embedding-0.6B~\cite{qwenemb} is the most aligned candidate by ridge $R^2$ and the second by
CKA, yet the SCOUT model trained against it retrieves \emph{worst} of all four ($61.37$ val mAP@10,
last row of \cref{tab:yencoder}). Adding it breaks the three lenses unevenly: mutual $k$NN overlap
stays moderately predictive ($\rho = 0.8$), CKA falls to $\rho = 0.4$, and ridge $R^2$ inverts
($\rho = -0.2$). With $n=4$ none of these reach significance ($p \geq 0.2$), so we report the
criterion as a training-free screening heuristic rather than a validated law, and the
neighborhood-overlap lens as the more robust of the three. The failure arises in the learned mapping
rather than in the target's static geometry. InfoNCE reaches its lowest training loss on this
target while generalizing worst over the $40$k gallery, and correcting for the pronounced
anisotropy of its embedding space does not repair the ordering. We also revisited the misaligned
PE-Core with a trainable projection on the frozen text encoder, the one component VL-JEPA adds and
SCOUT omits. The projection does not help, lowering accuracy further ($58.87$ val mAP@10) by
deepening the collapse of the predicted embeddings. CLIP-text is therefore the best frozen target we tested. EmbeddingGemma is
the second-best, and the ablation sequence of \cref{tab:ablation} builds on it.\looseness=-1

\begin{table}[tb]
  \centering
  \caption{Training-free X$\leftrightarrow$Y alignment orders the frozen text encoders identically
  to SCOUT val mAP within the CLIP/EmbeddingGemma/PE-Core family (Spearman $\rho=1.0$, all runs at
  $K=32$). The falsifier (last row) is the most aligned by ridge $R^2$ yet retrieves worst.}
  \label{tab:yencoder}
  \begin{tabular}{@{}lcccr@{}}
    \toprule
    Frozen text target $Y$ & kNN@10 & CKA & ridge $R^2$ & mAP@10 (\%) \\
    \midrule
    CLIP-ViT-L/14 text & 0.237 & 0.480 & 0.297 & \textbf{77.92} \\
    EmbeddingGemma-300M & 0.220 & 0.421 & 0.262 & 74.81 \\
    PE-Core-B-16 text & 0.196 & 0.384 & 0.230 & 67.57 \\
    \midrule
    Qwen3-Emb-0.6B {\footnotesize(falsifier)} & 0.218 & 0.444 & 0.308 & 61.37 \\
    \bottomrule
  \end{tabular}
\end{table}

\section{Two Precision-Targeted Levers}
\label{sec:levers}

On the final leaderboard, the top three teams lead us by $21$--$23$ points at rank 1 (R@1) but only $2$--$3$
at rank 10 (R@10). A gap this concentrated at the top is the signature of a \emph{precision} problem,
not a recall one. Offline, the fused bi-encoder pool already holds the ground truth in
its top-128 for $99.98\%$ of val queries, so deepening the rerank pool yields almost nothing, and the
remaining work is to promote the correct image to rank one. Both levers below start from the best
frozen single model ($89.32$ val mAP@10), which composes the CLIP-text target
(\cref{sec:alignment}) with the batch-128 recipe (\cref{sec:ablation}), and both act on R@1.

\subsection{ExPLoRA: Adapting the Frozen Video Encoder}
\label{sec:explora}
V-JEPA's self-supervised features were never shaped for
fine-grained person and action discrimination. Closing the R@1 gap therefore means reshaping
them by adapting the video encoder, carefully. A predictor is already trained on the frozen
features, so we are at the linear-probe stage of LP-FT~\cite{lpft}, where full fine-tuning would
distort the pretrained features. The appropriate second stage is ExPLoRA~\cite{explora}, which
unfreezes a small set of transformer blocks, applies LoRA to the rest, and tunes all normalization
layers. A relative-gradient-norm probe sets the unfreeze window. It finds a late-block semantic
peak plus a small early input-shift bump, whereas a naive sim-to-real prior would predict early
blocks only. We unfreeze the first two and last four blocks accordingly,
together with all LayerNorms and the patch embedding, and place LoRA ($r = 32$, $\alpha = 64$) on
the middle blocks. Learning rates are grouped: predictor $2 \times 10^{-5}$, unfrozen encoder
$5 \times 10^{-6}$ (gentle, OOD-preserving), LoRA $2 \times 10^{-4}$. We initialize from the trained
predictor and set the LoRA adapters to zero, so the encoder equals the frozen model at step
zero, in the spirit of WiSE-FT weight-space ensembling~\cite{wiseft}.

This is the first SCOUT configuration to train the video encoder, and it improves on the frozen
model, raising val mAP@10 from $89.32$ to $92.97$ ($+3.65$). The gain is predominantly precision.
R@10 barely moves ($+0.4$) while mAP jumps, which with one relevant image per query means the
ground truth itself moved toward rank one. On the leaderboard, we swap the adapted encoder into the
system's retriever pool (\cref{sec:system}) in place of its frozen twin, the CLIP-text model. The two
are redundant at Spearman $\rho = 0.64$, but the adapted encoder is decorrelated from the other
fusion members, $\rho = 0.22$--$0.34$. The swap adds $+0.51$ leaderboard mAP@10, concentrated at R@1
($+0.66$, with R@5 $+0.15$ and R@10 $+0.20$). The offline precision edge transferred to the
leaderboard as an R@1-led gain. Pushing the recipe harder (more
blocks, larger LoRA, higher encoder learning rate) saturates (\cref{sec:negatives}).

\subsection{Attribute-Decomposed VLM Reranking}
\label{sec:decomp}
The dominant term in our rerank blend is a vision-language model (VLM, here
Qwen3-VL-30B-A3B-Instruct~\cite{qwen3vl}) asked one holistic question per candidate (``does this
photo match the caption? yes/no''), read out as the log-probability of ``yes''. This term has a
measurable calibration failure, assigning $P(\mathrm{yes}) > 0.9$ to $6.4$ of every $20$
candidates on average. It discriminates only coarsely within its own top, exactly where R@1 is decided. We fix
it by decomposing the query along the three axes the task itself defines (\cref{sec:dataset}). One
VLM call scores appearance, action, and scene separately, each from $0$ to $100$. We combine the
three conjunctively by geometric mean, so a failure on any single aspect drives the combined score
down. This directly
targets the ``matches appearance, wrong action or scene'' confusion that a single yes/no blurs. The
decomposed score is a new signal, essentially uncorrelated with the holistic term (per-query
$\rho$ between $-0.09$ and $-0.005$).

We deploy the decomposed score added to the holistic term, half and half within the rerank slot,
rather than as a replacement. Fully replacing the holistic term changes
rank one for $19.1\%$ of test queries. Most are benign tie-breaks, but a quarter are blind overrides
that promote a holistic-rejected candidate, and those cannot be sized offline. Keeping half of the
holistic term preserves the calibration fix while damping those overrides. Added this way, the
decomposed reranker improves the leaderboard score by $+1.23$ mAP@10 / $+1.57$ R@1, to $84.18$
mAP@10. Our best submission, $84.25$ on the final
leaderboard, adds only a fusion-member swap on top. The negative result of
\cref{sec:negatives} confirms why, the holistic magnitude carrying top-1 evidence that a
decorrelated signal must be layered on, not substituted for.

\begin{table}[tb]
  \centering
  \caption{The two precision levers. The \emph{val} column is the standalone single model, while
  the \emph{leaderboard} columns are the full system of \cref{sec:systemeval} with the row's
  component included, scored post-close on the full test set. The last row is the winning team,
  for scale.\looseness=-1}
  \label{tab:levers}
  \begin{tabular}{@{}lcccc@{}}
    \toprule
     & val (model) & \multicolumn{3}{c@{}}{leaderboard (system)} \\
    \cmidrule(l){3-5}
    Lever & mAP@10 & mAP@10 & R@1 & R@10 \\
    \midrule
    Frozen base model (CLIP-Y) & 89.32 & 82.43 & 73.05 & 96.51 \\
    + ExPLoRA video-encoder adaptation & \textbf{92.97} & 82.94 & 73.71 & 96.71 \\
    + attribute-decomposed VLM rerank & n/a & 84.18 & 75.28 & \textbf{96.92} \\
    + fusion-member swap (best submission) & n/a & \textbf{84.25} & \textbf{75.63} & 96.66 \\
    \midrule
    Final leaderboard best (method undisclosed) & n/a & 99.30 & 98.74 & 100.00 \\
    \bottomrule
  \end{tabular}
\end{table}

\section{System and Sim-to-Real Evaluation}
\label{sec:systemeval}

\subsection{The Ensemble System}
\label{sec:system}
\begin{figure}[tb]
  \centering
  \includegraphics[width=\linewidth]{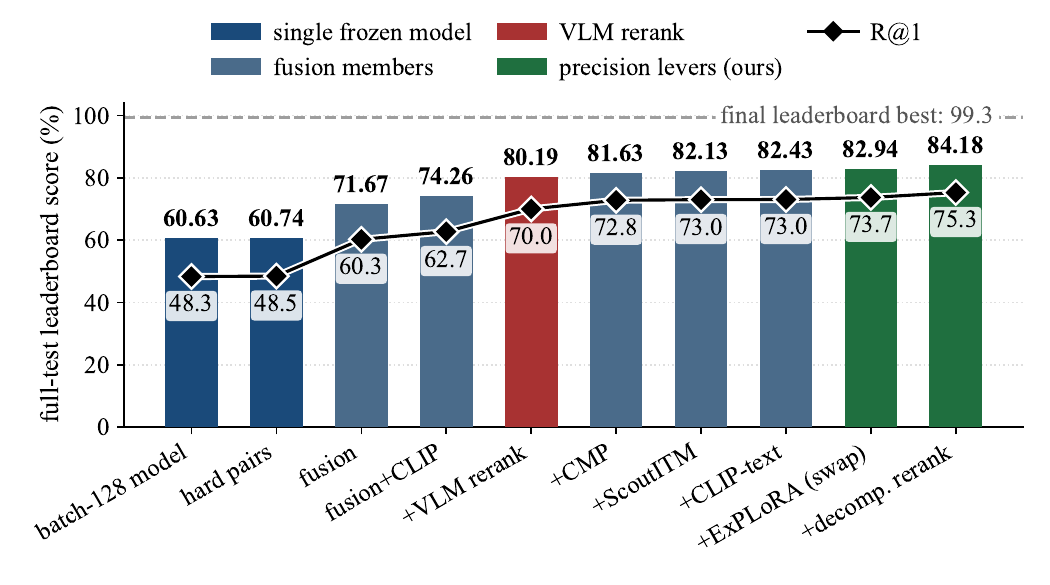}
  \caption{Leaderboard progression (full-test mAP@10 / R@1) from a single frozen model ($60.63$)
  to the full system ($84.18$, best submission $84.25$), against the final leaderboard
  best.}
  \label{fig:board}
\end{figure}

On top of the trained model we build a retrieve-fuse-rerank ensemble for the leaderboard.
It adds no further training, so leaderboard numbers are \emph{system} scores rather than
single-model ones. Its bi-encoders and the ScoutITM reranker operate
on \emph{frozen} features, while the VLM reranker reads raw images with a frozen off-the-shelf model.
Submitted alone, the best frozen single model scores $60.63$ leaderboard mAP@10 against $86.57$ on
val (\cref{sec:calib} analyzes the drop). The ensemble closes that gap in three moves,
traced submission by submission in \cref{fig:board}: fusing decorrelated retrievers recovers
$+13.6$ points, the three rerank signals below add a further $+7.7$, and retriever-pool upgrades,
including the ExPLoRA swap of \cref{sec:explora} and the final member swap of the best submission
($84.25$), the remaining $+2.3$.\looseness=-1

\textbf{Retrieve.} We run a set of architecturally decorrelated bi-encoder retrievers, each producing
a per-query ranking of the gallery. The set spans four SCOUT variants (the ExPLoRA-adapted video
encoder of \cref{sec:explora} and three frozen-encoder checkpoints), the frozen CLIP~\cite{clip} and
SigLIP2~\cite{siglip2} dual encoders, and the dataset authors' CMP~\cite{cmp} dual-encoder scores.

\textbf{Fuse.} We combine the retrievers by CombSUM~\cite{combsum}. For each query we
min-max-normalize every retriever's similarities into $[0,1]$ and sum them, keeping the top-$20$ as a
short list. We fuse at the score level because the members differ in \emph{quality}, and summing
normalized scores lets a strong member outvote a weak one. A pure rank
re-blend (reciprocal-rank fusion~\cite{rrf}) discards these magnitudes and regressed on the
leaderboard (\cref{sec:calib}).

\textbf{Rerank.} The short list is re-scored by a per-query min-max blend of three signals: the
fused score itself (weight $0.2$), the Qwen3-VL cross-encoder combining its holistic match
probability with the attribute-decomposed refinement of \cref{sec:decomp} (weight $0.6$), and
ScoutITM (\cref{sec:itm}, weight $0.2$). The cross-encoder supplies the query-conditioned
cross-attention a bi-encoder pool structurally lacks, and is the single largest gain, raising
leaderboard R@1 from $62.7$ for the fused pool to $70.0$. Each added term follows the rule of \cref{sec:decomp,sec:calib},
\emph{decorrelated} from the others and \emph{added} rather than substituted for the dominant
signal.

\subsection{ScoutITM: A Frozen-Feature Cross-Encoder Reranker}
\label{sec:itm}
ScoutITM, the third rerank term, is an image-text matching (ITM) head that scores an (image, caption)
pair \emph{jointly}, attending across modalities before a single score. Both encoders stay frozen, exactly as in SCOUT.
V-JEPA emits patch tokens and EmbeddingGemma emits text-token states. A shallow \emph{single-stream}
transformer ($29.73$M trainable parameters) attends over the concatenated token sequence, reading
out a scalar match probability through a binary head. It is trained by binary cross-entropy on matched
pairs against identity-based and short-list-mined negatives from a frozen, already-trained SCOUT
retriever, so it learns to separate exactly the confusable candidates a bi-encoder ranks near the top.\looseness=-1

It ranks differently from both the bi-encoder pool and the VLM (per-query $\rho = 0.21$), a
genuinely decorrelated term rather than a redundant refinement, and adding it to the two-term blend
of fused score and VLM improves the leaderboard score by $+0.50$ mAP@10.
Unlike the dataset authors' fully fine-tuned cross-encoder, ScoutITM
trains only this head over cached frozen features in about 20 GPU-hours, and it can be evaluated on
our held-out split without leakage (\cref{sec:calib}).

\begin{figure}[tb]
  \centering
  \includegraphics[width=0.70\linewidth]{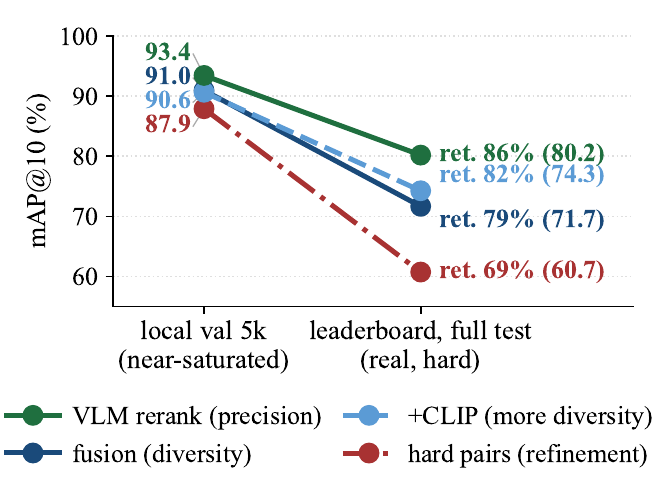}
  \caption{Local (val) against leaderboard (full-test) mAP@10 by intervention type, with each
  line's retention (leaderboard over val). The dashed +CLIP line starts below fusion on val yet
  lands above it on the leaderboard, the sign flip discussed in the text.}
  \label{fig:calib}
\end{figure}

\subsection{Local vs.\ Public Calibration}
\label{sec:calib}
Every intervention was measured on held-out val before being used for a leaderboard submission, and
the question is how much of a val gain survives there. We
quote all deltas in \emph{points} of mAP@10. The
central finding (\cref{fig:calib}) is that this calibration \emph{reverses by intervention type}
on the interventions we tested. A single-model retention factor of $0.70$ appeared stable
($86.57$ val, $60.63$ leaderboard) but did not hold. The gains of \emph{diversity} interventions grow on the
hard real test. Late fusion gained $+3.0$ points over the best single model on val but $+10.9$ on
the leaderboard ($\times 3.6$).
Adding a decorrelated frozen-CLIP member to that fusion \emph{lost} $0.3$ points on val yet gained
$2.6$ on the
leaderboard, so val had the wrong sign. A diversity gain scales with how much error remains to fix,
and the real test has far more (the fused pool's R@1 is $60.3$ on the leaderboard against a
near-saturated $85.6$ on val).\looseness=-1

Conversely, the gains of reorder-only \emph{refinements} shrink or reverse. A rank-fusion re-blend
of the cached
VLM scores gained $+3.0$ points over the standard min-max blend on a harder validation split built
from the pre-mined hard pairs,
yet scored $-0.52$ against the same blend on the leaderboard. A single training-free
statistic predicts what transfers, the per-query rank decorrelation (Spearman $\rho$) of a candidate
signal from the existing members. Every leaderboard gain we recorded (fusion, the VLM rerank, ScoutITM,
ExPLoRA, the decomposed reranker) was a decorrelated \emph{add} or a precision lever, while reorder
refinements and redundant adds regressed or produced no gain. We offer this as a heuristic, the
evidence being a handful of interventions on one benchmark, and $\rho$ predicts direction, not
magnitude.\looseness=-1

In-distribution proxies \emph{bracket} the leaderboard, the near-saturated val split
under-predicting and a $200$k-distractor proxy over-predicting, so under a 20-submission cap we
used them only for direction and decorrelation, never for the magnitude of a gain. CMP, fine-tuned
on the very corpus our splits draw from, scored a leaked $98$ val mAP@10 and can only be assessed
on the leaderboard. That cap also meant we never submitted CMP, CLIP, or SigLIP2 standalone, or
resubmitted the final system with any removed, reserving it for interventions with an expected
gain over diagnostic ablations before the challenge closed. CMP added $+1.44$ mAP@10 joining an
earlier, smaller retriever pool (\cref{fig:board}), not a controlled ablation of the final system,
and CLIP added $+2.59$ mAP@10 / $+2.48$ R@1 joining a pool that already held SigLIP2, which has no
equivalent delta of its own. CMP's own paper~\cite{cmp} reports $91.66$ mAP@10, but its own test
images substitute for the gallery ($1{,}978$ against our $36{,}773$), so the two numbers are not on
the same task. The in-challenge subset was itself noisy, full-test rescoring moving submissions by
up to a point and shrinking the ScoutITM gain from $+1.16$ to $+0.50$.\looseness=-1

\subsection{Negative Results and Limitations}
\label{sec:negatives}
Where the gains stop is as informative as the gains. On val, a larger \emph{frozen}
video encoder reduces accuracy. The
raw V-JEPA ViT-g scores $81.86$ against our distilled ViT-L's $89.32$. Also on val, a stronger
ExPLoRA recipe saturates ($92.82$ against the conservative recipe's $92.97$), so video-encoder
adaptation does
not scale by strengthening the recipe. On the leaderboard, fusion \emph{membership} saturated after
the video-encoder lever, three further member changes moving the score by at most $+0.30$ mAP@10
each (exact ties on the in-challenge subset), so decorrelation predicts diversity, not remaining
gain. The add-don't-replace rule of \cref{sec:decomp} has a sharp negative counterpart.
\emph{Replacing} the holistic VLM with the decomposed score inside its saturated cluster reduced
the leaderboard score by $4.28$ points relative to the half-and-half addition of \cref{sec:decomp},
since the holistic $P(\mathrm{yes})$ magnitude carries top-1 signal on
the OOD test set, and discarding it (as the rank re-blend did) reverses the gain (\cref{fig:board}).
\looseness=-1

\section{Conclusion}
\label{sec:conclusion}

SCOUT needs no fine-tuned cross-encoder for strong sim-to-real retrieval, though the remaining gap
to the top teams is top-rank precision, methods undisclosed. The alignment criterion and
calibration study both remain heuristics (one encoder family, one benchmark we also used to select
submissions, not an untouched holdout). A video encoder
untrained on text maps well enough into a frozen caption space to compete, worth wider
study.\looseness=-1

\section*{Acknowledgements}
We thank Embia Lab, Universit\'e de Moncton, for the compute infrastructure that made this work
possible.

% ---- Bibliography ----
\bibliographystyle{splncs04}
\bibliography{refs}

\end{document}